\documentclass{llncs}
\usepackage{amssymb}
\usepackage{amsmath}
\usepackage{multirow}
\usepackage{graphicx}

\begin{document}
\frontmatter          % for the preliminaries
\mainmatter              % start of your contributions
\title{Conv2Warp: An unsupervised deformable image registration with continuous convolution and warping}
\author{Sharib Ali and Jens Rittscher}
\institute{Department of Engineering Science, University of Oxford, Institute of Biomedical Engineering, Oxford, UK\\  \email{\{sharib.ali, jens.rittscher\}@eng.ox.ac.uk}}
\authorrunning{Ali and Rittscher}
\titlerunning{Conv2Warp: An unsupervised deformable image registration}
\maketitle           
\begin{abstract}
Recent successes in deep learning based deformable image registration (DIR) methods have demonstrated that complex deformation can be learnt directly from data while reducing computation time when compared to traditional methods. However, the reliance on fully linear convolutional layers imposes a uniform sampling of pixel/voxel locations which ultimately limits their performance. To address this problem, we propose a novel approach of learning a continuous warp of the source image. Here, the required deformation vector fields are obtained from a concatenated linear and non-linear convolution layers and a learnable bicubic Catmull-Rom spline resampler. This allows to compute smooth deformation field and more accurate alignment compared to using only linear convolutions and linear resampling. In addition, the continuous warping technique penalizes disagreements that are due to topological changes. Our experiments demonstrate that this approach manages to capture large non-linear deformations and minimizes the propagation of interpolation errors. While improving accuracy the method is computationally efficient. We present comparative results on a range of public 4D CT lung (POPI) and brain datasets (CUMC12, MGH10).
\end{abstract}
\section{Introduction}
\label{sec:intro}
Image registration is required in many medical imaging applications - from multi-modal data fusion to inter- and intra-patient comparisons. Rigid or affine registration methods can be used to align source image to its target image by computing a simple transformation matrix, i.e., optimization of only few parameters. However, such an alignment does not model changes due to organ deformation, patient weight loss, or tumour shrinkage. In order to tackle these changes non-rigid or deformable image registration (DIR) methods are advised. Traditional DIR methods are usually mathematically complex to model and therefore require the optimization of a very larger number of parameters which makes them computationally expensive. Recently, deep learning based methods have shown tremendous success in tackling such complex problems in image registration. This is mainly due to two reasons: 1) By learning the model parameters of the neural network through optimizing a loss function directly from the data a wide range of image features are considered. 2) These methods make very efficient use of hardware acceleration as a result they require significantly less time than traditional methods.

Recently, supervised as well as unsupervised learning techniques for medical image registration have been proposed. Supervised techniques~\cite{Krebs:MICCAI17} rely on ground truth deformation vector fields (DVFs) computed by traditional methods. Here, a neural network is only used as a regressor for approximating traditional methods. This restricts their learning process and such training requires a large number of DVFs which are both time consuming to generate and carries the risk of mitigating unreliable deformations. On the other hand, unsupervised techniques~\cite{Li:ISBI18,Guha:TMI2019,Bob:MedIA2019} can learn to predict DVFs without requiring ground truth deformations. Linear convolutional neural networks (ConvNets) were used in~\cite{Li:ISBI18,Guha:TMI2019,Bob:MedIA2019}. Guha et al.~\cite{Guha:TMI2019} and Li et al.~\cite{Li:ISBI18} used linear sampling techniques for upscaling the obtained DVFs. We argue that such models do not approximate complex and non-linear local deformations with a sufficient accuracy. These methods were evaluated on brain datasets where local deformations are limited. While,~\cite{Bob:MedIA2019} proposed to use a B-spline as transformation model and interpolation method for the predicted DVFs  and presented results on more complex cardiac cine MRI and lung datasets. However, in~\cite{Bob:MedIA2019} only linear ConvNets were used. Despite of good local support, B-splines can lead to larger interpolation errors as they do not pass through data points (see {Suppl. Mat. Section~1} for details). 

With this paper, we introduce an unsupervised approach for deformable image registration that is capable of learning complex deformations and is able to compute smooth, accurate and plausible DVFs. We propose to: 1) Relax the fixed geometric constrain imposed by traditional convolutional filters by adding a series of deformable convolutional filters~\cite{Dai:ICCV17} combined with linear convolutional filters which allows to capture complex features. 2) Apply a learnt bicubic Catmull-Rom spline resampler to minimize error in DVF resampling, 3) Aggregate large deformations using a multiple scale warping strategy. 4) Finally, to impose further smoothness of the DVF using a $L_2$-norm regularization. We show that interpolation technique influence the learning phenomena. We also illustrate deformable convolutions as presented in this work through experiments. We have evaluated our model on publicly available lung CT and brain MRI datasets.
\begin{figure}[t!]
\centering
\includegraphics[scale=0.24]{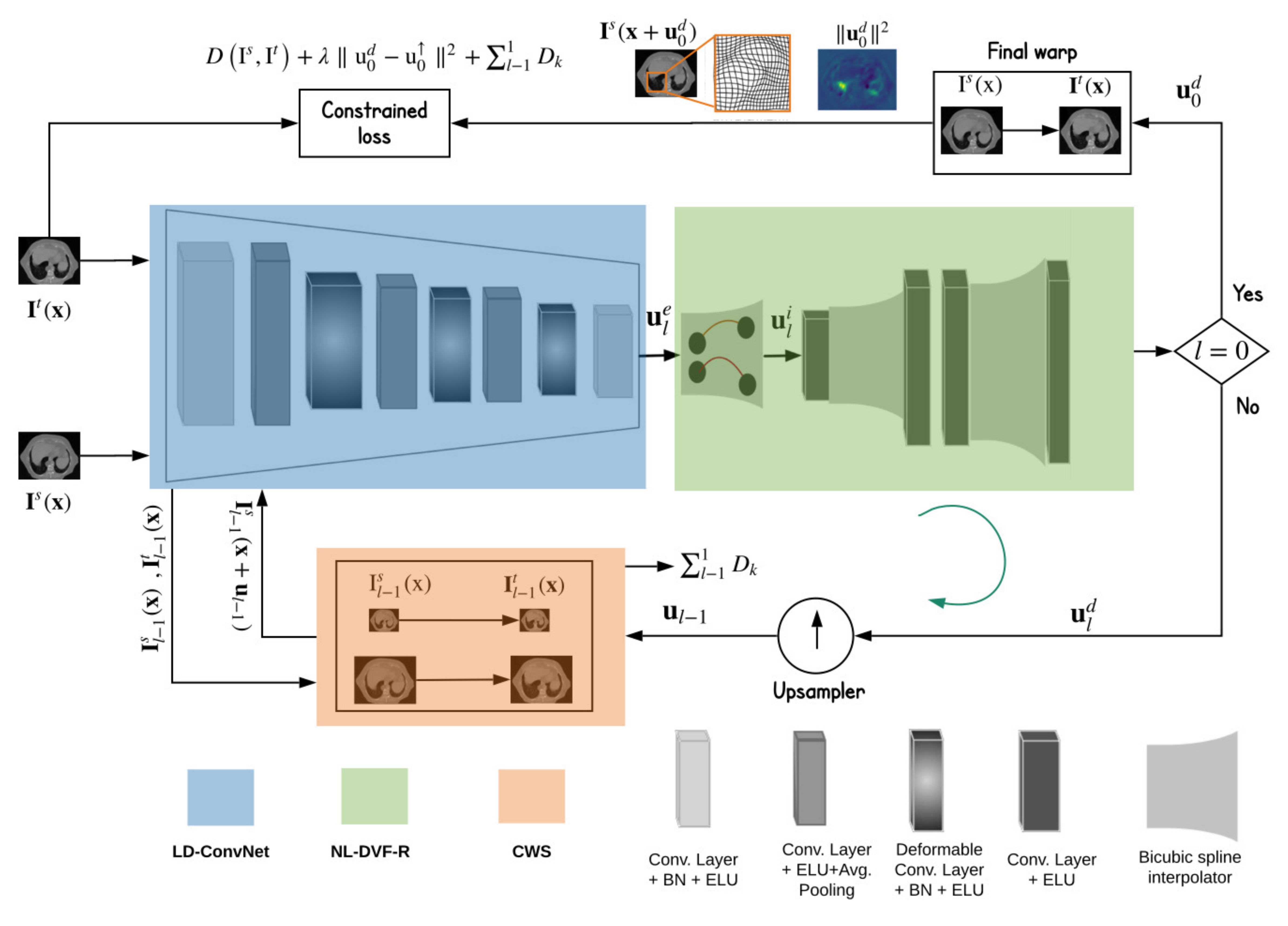}
\caption{\textbf{Conv2Warp model.} Three main components are presented as colored blocks:  Blue: linear and deformable convolution network (LD-ConvNet), Green: non-linear DVF resampler network (NL-DVF-R) and Orange:  continuous warping stage (CWS).  {\label{fig:block}}}
\end{figure}
\section{Method}
\label{sec:method}
As motivated, we now introduce the key elements to our Conv2Warp model (each highlighted in Fig.~\ref{fig:block}), and describe our constrained loss function. 
\paragraph{\textbf{Linear and Deformable Convolutional Network (LD-ConvNet)}} LD-ConvNet (blue block in Fig.~\ref{fig:block}) consists of 5 linear convolutions and 3 deformable convolution layers. Each convolution layer is combined with a batch normalization and ELU activation function. After the second convolution layer an average pooling layer with downsampling factor of 0.5 is applied. Throughout we use a sequential ConvNet of (input, output) channels as follows: $(2, 64)$,  $(64, 64)$, $(64, 64)$, $(64,32)$, $(32, 32)$, $(32,16)$, $(16,16)$, $(16,2)$ with kernel size $3$ and stride 1. Two images $I^s$ ({\it{source}}) and $I^t$ ({\it{target}}) are concatenated first and set as an input to the first layer of the Conv2Warp model. 
\paragraph{\textbf{Non-linear DVF Resampler (NL-DVF-R)}}
{NL-DVF-R} (green block in Fig.~\ref{fig:block}) consists of a sequential bicubic Catmull-Rom spline resampler and convolutional filter for resampling the obtained DVF from the LD-ConvNet. Since, this is integrated in our learning model, complex non-linear deformations  learnt by the LD-ConvNet are guaranteed to have smooth deformation fields and the interpolation error  is minimal. 

Catmull-Rom spline consist of 4 basis functions with local support and are $C_1$ continuous and differentiable (see {Suppl. Mat. Section~1}). These properties makes them smoother compared to standard linear interpolation techniques. Qualitative comparison of Catmull-Rom spline with linear resampling technique is provided in  Fig. 3 of Suppl. Mat.

\begin{figure}[t!]
\centering
\includegraphics[trim=0.0cm 0.2cm 0.15cm 0.05cm, clip=true,scale=0.28]{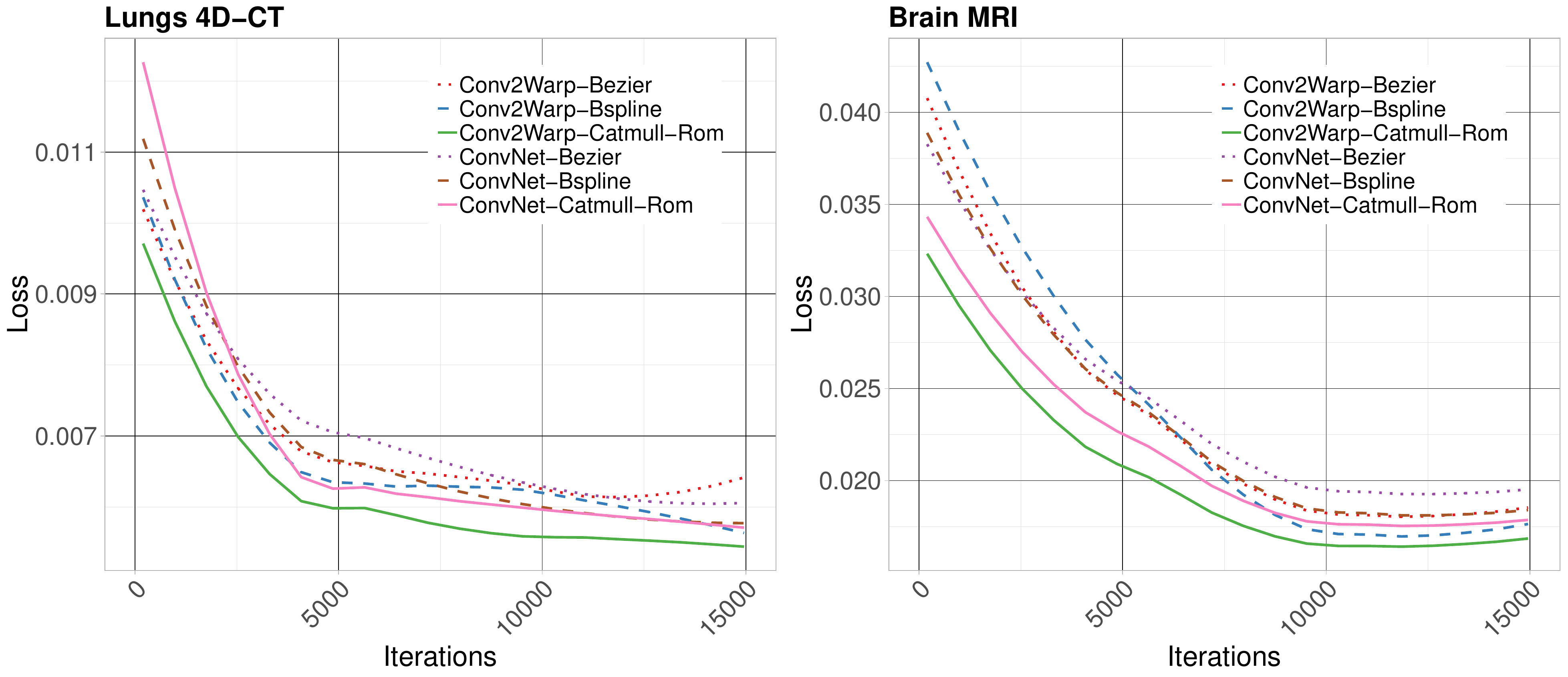}
\caption{\textbf{Training losses:} Effect of interpolation methods on linear convolution model (ConvNet) and our proposed non-linear Conv2Warp model. Conv2warp has better convergence than the linear model for every interpolation technique, while the best with the Catmull-Rom spline interpolation. \label{fig:training_loss}}
\end{figure}

Comparison of Catmull-Rom spline with other spline-based resamplers (including B-spline) is presented in Fig.~\ref{fig:training_loss}. It can be observed that Catmull-Rom spline has the lowest training losses for both lung and brain datasets. 
\paragraph{\textbf{Continuous Warping Stage (CWS)}}
CWS (orange block in Fig.~\ref{fig:block}) warps the deformation field obtained at $l$ pyramid levels in a continuous fashion. Here $l = 4$ for our 2D case with image size $256\times 256$ and $l=2$ for our 3D case with patch volume size $64\times64\times64$. Note that $l$ pyramids are constructed prior to feeding images into the LD-ConvNet. 

The CWS employs warping of the next level source image $I^s_{l-1}$ with the computed DVF from previous coarse level LD-ConvNet, i.e., $\textbf{u}_{l-1}=\textbf{u}^{d\uparrow}_{l}$. This process is repeated until the final level, i.e., $l =0$. At each level the losses $\mathit{D}\left({I}^s_l,{I}^t_l\right)$ between the warped image $I^s_{l-1}\left(\textbf{x}+\textbf{u}_{l-1}\right)$ and the target are summed and transferred to the final constrained loss function detailed in the Section~\ref{subsec:loss}.
\paragraph{\textbf{Loss function}}{\label{subsec:loss}}
% Data-term
A multi-modal normalized cross-correlation (NCC) metric is used as data term $\mathit{D}$ in our loss function. We propose to use a derived sum-of-squared difference (SSD) which guarantees maximization of NCC metric~\cite{DruleaNedevschi13:TIP} and is written as
$\mathit{D}\left({I}^s,{I}^t\right) =\tfrac{1}{2N}\sum \left( \tfrac{ \mid I^s(\textbf{x})-\mu^s \mid}{{\sqrt{\sigma_s^2+\epsilon^2}}}- \tfrac{ \mid I^t(\textbf{x})-\mu^t \mid}{\sqrt{\sigma_t^2+\epsilon^2}}\right)^2,$
with  mean $\mu$, standard deviation $\sigma$, the total number of pixels $N$ and $\epsilon=10^{-3}$ (used to avoid division by zero).
{For} obtaining a smooth deformation vector field we apply an $L_2$-norm of the difference between the computed $\textbf{u}^d_{0}$ from the NL-DVF-R block and $\textbf{u}_0^{\uparrow }$ which is the previous upsampled input DVF to CWS block (see Fig.~\ref{fig:block}). Losses computed during $l$ pyramid levels in our CWS block $ \sum_{l-1}^{1} \mathit{D}_k$ are aggregated to the final loss. The resulting equation for the backward propagation is:
$\mathcal{L}= \mathit{D}\left({I}^s,{I}^t\right)+\lambda  \left[ \parallel\textbf{u}_0^d-\textbf{u}_0^{\uparrow}\parallel^2\right] _{0}^{0.25}+ \sum_{l-1}^{1} \mathit{D}_k,$
where $\lambda$ is the trade-off between the regularization and the data-term. In our experiments  it is set to  0.001. We restrict the regularization to the interval $[0, 0.25]$ to prevent over smoothing problem during training. 
Omitting this constraints results in failures which are documented in Fig. 2 of Suppl. Mat. (see plot for Conv2Warp-FullReg). The loss saturates after 6000 iterations.  However, regularised models with the interval continue to minimise the loss function. Our constrained loss is optimized using Adam optimizer with learning rate $10^{-3}$.
%Minimization of the formulated SSD will lead to maximization of NCC, i.e., $\lim_{\mathit{D}(I^s, I^t)\to0} NCC(I^s, I^t)= 1$. .~\ref{fig:training_loss}
\section{Experiments and Results}
\label{sec:exp}
\subsection{Datasets and training}
\textbf{4D CT.} DIR-LAB~\cite{DIRLAB:PhyMed09} thoracic image dataset for DIR that inlcudes inspiratory and expiratory breath-hold CT image pairs were used for training. Although large permutations are possible, only breathing cycles 00-50, 10-80 and 30-90 were considered from 10 different sets (a total of 120 volumes). As a consequence we avoid an imbalance in the training dataset as most of the breathing cycles did not present strong deformations.
For testing the POPI~\cite{POPI11} lung dataset was used. The provided segmentation for air, body and lungs were used for evaluation.

\noindent
\textbf{MRI.} The LBPA40~\cite{LBPA40} dataset were used to train the model. We performed a random combination of these brain pairs and used nearly 300 volumes for our training.  We have used MGH10 and CUMC12~\cite{Klein:NI09} datasets for which masks are available for evaluation. 300 epochs with batch size of 4 were used for training performed on a 16GB NVIDIA Tesla P100 for nearly 48 hours. 
\begin{table}[t!]
\centering
\small{
\begin{tabular}{llllllllllllllllllllll}
\hline
\multirow{2}{*}{{\bf 4D-CT}} & \multicolumn{15}{c}{{\bf Methods}}   \\                                                
\cline{2-9} 
& \multicolumn{3}{c}{{\bf Pre-align.}} & \multicolumn{3}{c}{{\bf Demons}} & \multicolumn{3}{c}{{\bf ANTS-SyN}}& \multicolumn{3}{c}{{\bf SE}} & \multicolumn{3}{c}{{\bf ConvNet}} & \multicolumn{3}{c}{{\bf Conv2Warp}}   \\
                        \hline
00-50        & \multicolumn{3}{c}{0.83(0.76)}               & \multicolumn{3}{c}{0.87(0.83)} & \multicolumn{3}{c}{0.87(0.82)}    & \multicolumn{3}{c}{\textbf{0.89(0.85)}}   & \multicolumn{3}{c}{0.87(0.82)} & \multicolumn{3}{c}{\textbf{0.89}(0.83)}  \\
00-70            & \multicolumn{3}{c}{0.86(0.79)}           & \multicolumn{3}{c}{0.90(0.85)} & \multicolumn{3}{c}{0.89(0.84)}    & \multicolumn{3}{c}{\textbf{0.92(0.88)}}   & \multicolumn{3}{c}{0.90(0.84)} & \multicolumn{3}{c}{0.91(0.85)}   \\
10-50              & \multicolumn{3}{c}{0.82(0.75)}         & \multicolumn{3}{c}{0.86(0.82)} & \multicolumn{3}{c}{0.86(0.81)}    & \multicolumn{3}{c}\textbf{\textbf{0.88(0.85)}}   & \multicolumn{3}{c}{0.87(0.81)} & \multicolumn{3}{c}{\textbf{0.88}(0.83)}    \\
10-70            & \multicolumn{3}{c}{0.85(0.78)}           & \multicolumn{3}{c}{0.88(0.83)} & \multicolumn{3}{c}{0.87(0.82)}    & \multicolumn{3}{c}{0.88(0.84)}   & \multicolumn{3}{c}{0.89(0.83)} & \multicolumn{3}{c}{\textbf{0.90(0.84)}}    \\
 20-50             & \multicolumn{3}{c}{0.84(0.77)}          & \multicolumn{3}{c}{0.87(0.83)} & \multicolumn{3}{c}{0.86(0.82)}    & \multicolumn{3}{c}{\textbf{0.91(0.86)}}   & \multicolumn{3}{c}{0.87(0.81)} & \multicolumn{3}{c}{0.89(0.85)}    \\
 20-70            & \multicolumn{3}{c}{0.86(0.79)}           & \multicolumn{3}{c}{0.89(0.84)} & \multicolumn{3}{c}{0.88(0.83)}    & \multicolumn{3}{c}{0.88\textbf{(0.84)}}   & \multicolumn{3}{c}{0.89(0.83)} & \multicolumn{3}{c}{\textbf{0.90(0.84)}}    \\
 30-50          & \multicolumn{3}{c}{0.87(0.82)}             & \multicolumn{3}{c}{{0.90}(0.86)} & \multicolumn{3}{c}{0.89(0.85)}    & \multicolumn{3}{c}{\textbf{0.91(0.87)}}   & \multicolumn{3}{c}{0.88(0.83)} & \multicolumn{3}{c}{{0.90}(0.86)}    \\
 \hline
 \hline
Mean            & \multicolumn{3}{c}{0.84(0.78)}         & \multicolumn{3}{c}{0.88(0.83)} & \multicolumn{3}{c}{0.87(0.82)}    & \multicolumn{3}{c}{\textbf{0.90(0.85)}}   & \multicolumn{3}{c}{0.88(0.82)} & \multicolumn{3}{c}{\textbf{0.90}(0.84)}   \\
$\bar{t}$ (in s.)     &		\multicolumn{3}{c}{ - }          & \multicolumn{3}{c}{23.51}     & \multicolumn{3}{c}{216.45}    & \multicolumn{3}{c}{416.78}   & \multicolumn{3}{c}{\textbf{2.94}} & \multicolumn{3}{c}{\textbf{2.94}} \\
\hline
\end{tabular}
}
\caption{{\textbf{Evaluation on lung 4D CT dataset (POPI~\cite{POPI11}).} Dice(Jaccard) coefficients for 7 different volume pairs with larger inhalation and exhalation cyles. Grid-parameter search was done for all methods and only best performance is reported.~\label{tab:POPI} }}
\end{table}
\begin{table}[t!]
\centering
\small{
\begin{tabular}{ll|l|l|lll|l|l}
\hline
\multirow{3}{*}{{\bf Methods}} & \multicolumn{4}{c}{{\bf Dataset}} &  \multicolumn{1}{|c}{{\bf Avg. }}  \\                                                
\cline{2-5} 
& \multicolumn{2}{c|}{{\bf MGH10}} & \multicolumn{2}{c}{{\bf CUMC12}} & \multicolumn{1}{|c}{{\bf time}}
  \\
  & \multicolumn{1}{c|}{\bf $\mu_{dice}$} & \multicolumn{1}{c|}{{\bf $\mu_{jaccard}$}} & \multicolumn{1}{c}{{\bf $\mu_{dice}$}} & \multicolumn{1}{|c|}{{\bf $\mu_{jaccard}$}} & \multicolumn{1}{|c}{{\bf $\bar{t}$}}
  \\
 \hline
Rigid & 0.92 $\pm$ {0.011} & 0.86 $\pm$ 0.019        & 0.90 $\pm$ {0.030} & 0.82 $\pm$ 0.050  &  \multicolumn{1}{|c}{-}\\
\hline
SimpleElastix (SE)       &  \textbf{0.95} $\pm$ \textbf{0.003}   & \textbf{0.90} $\pm$ \textbf{0.005} & 0.95 $\pm$ \textbf{0.002}   & 0.91 $\pm$ \textbf{0.002}  &  \multicolumn{1}{|c}{1008.2}\\
ANTS (SyN-CC)       &  0.94 $\pm$ 0.007  & \textbf{0.90} $\pm$ 0.011 & 0.95 $\pm$ 0.010   & 0.90 $\pm$ 0.017   &  \multicolumn{1}{|c}{246.4}\\
\hline
ConvNet  & 0.93 $\pm$ {0.008}      & 0.88 $\pm$ 0.014      & 0.95 $\pm${0.007}      & 0.91 $\pm$ 0.012   &  \multicolumn{1}{|c}{\textbf{2.9}}\\
Conv2Warp & \textbf{0.95} $\pm$ 0.007   & \textbf{0.90} $\pm$ 0.012       &  \textbf{0.97}$\pm$ {0.005}   & \textbf{0.93} $\pm$ {0.008}  &  \multicolumn{1}{|c}{\textbf{2.9}}\\
  \hline \\
\end{tabular}
}
\caption{\textbf{Evaluation on T1-weighted MRI datasets.} Mean Dice and Jaccard coefficients  for two datasets with the state-of-the-art DIR methods and Conv2Warp (checkpoints trained on LBPA40 dataset) are shown. Only rigid registration was performed for pre-alignment prior to applying deformable registration.~\label{tab:BRAIN} }
\end{table}
\begin{figure}[t!]
\centering
%lung
\begin{minipage}[b]{0.135\linewidth}
\includegraphics[scale=0.19]{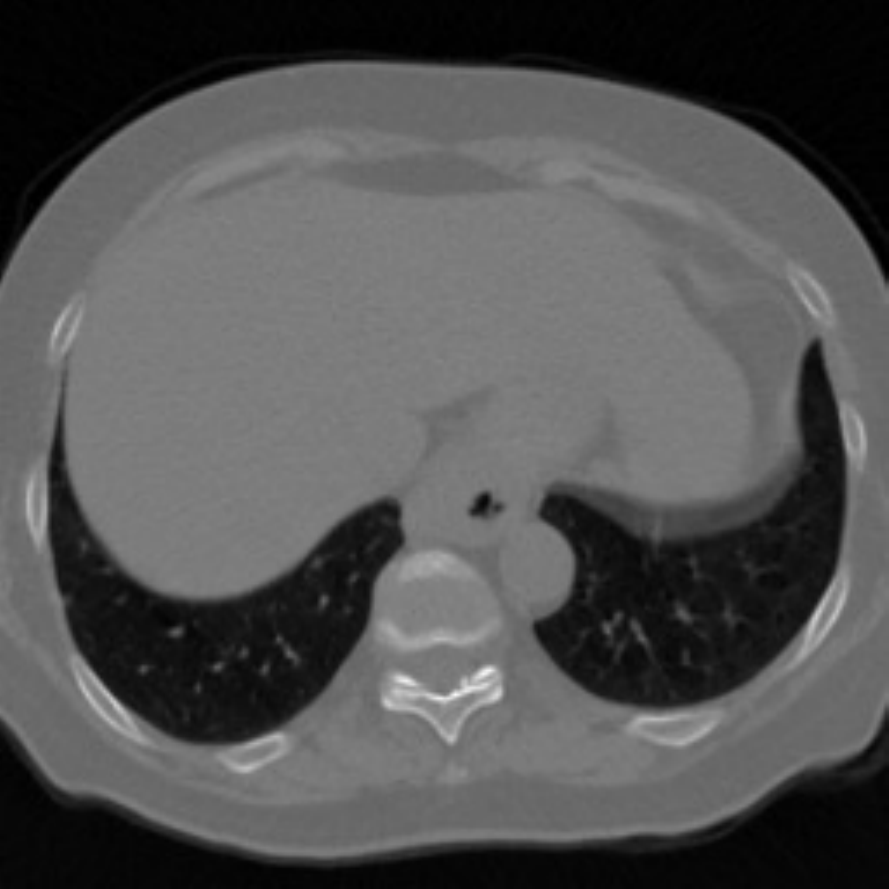}
\end{minipage}
\begin{minipage}[b]{0.135\linewidth}
\includegraphics[scale=0.19]{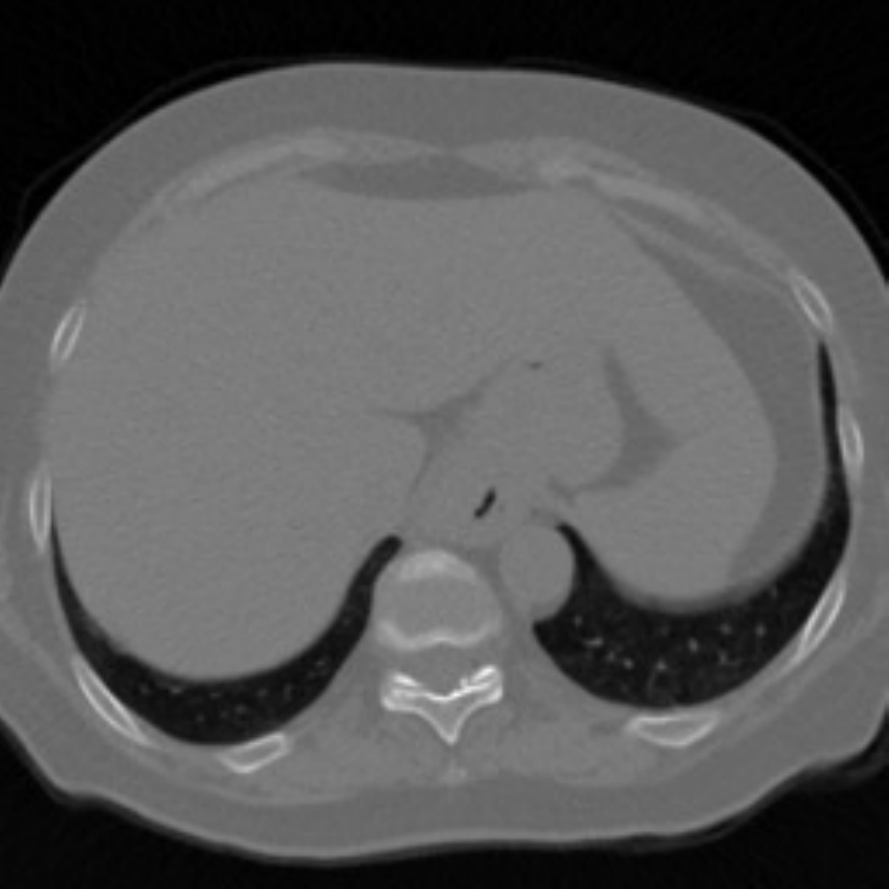}
\end{minipage}
\begin{minipage}[b]{0.135\linewidth}
\includegraphics[scale=0.19]{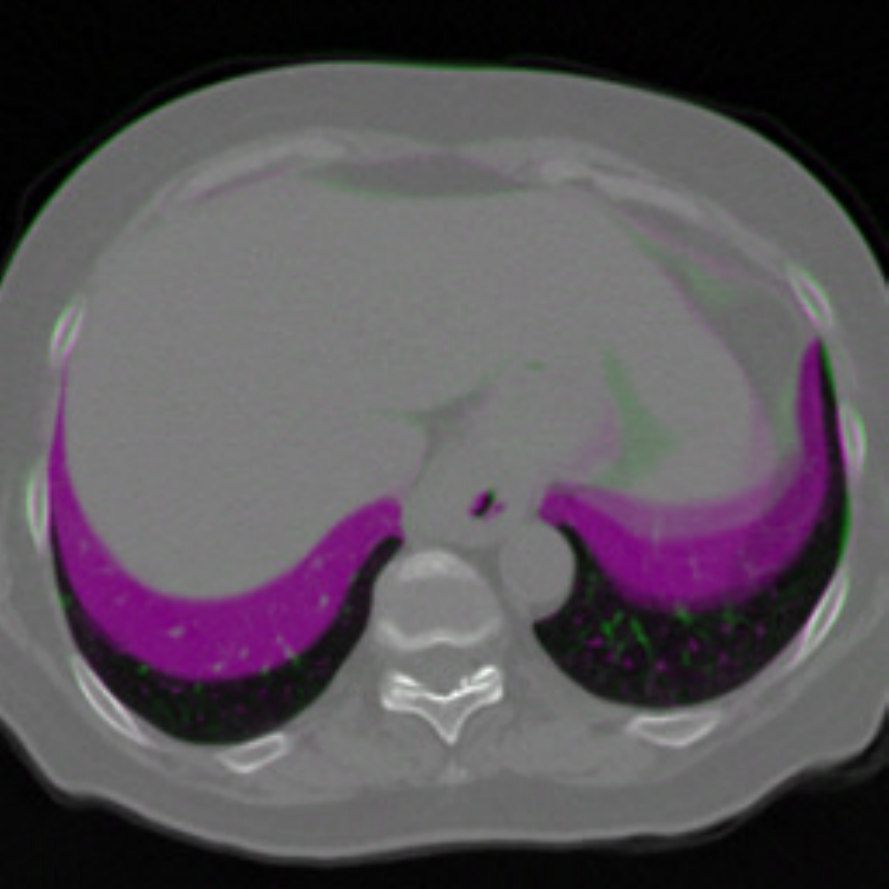}
\end{minipage}
\begin{minipage}[b]{0.135\linewidth}
\includegraphics[scale=0.19]{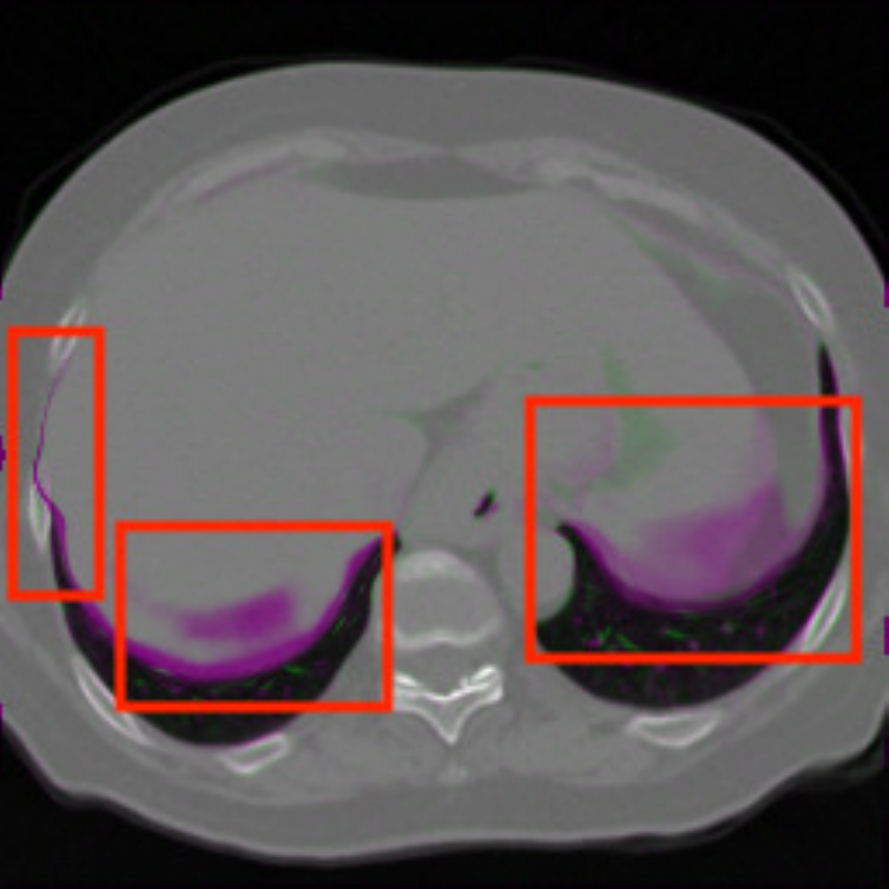}
\end{minipage}
\begin{minipage}[b]{0.135\linewidth}
\includegraphics[scale=0.19]{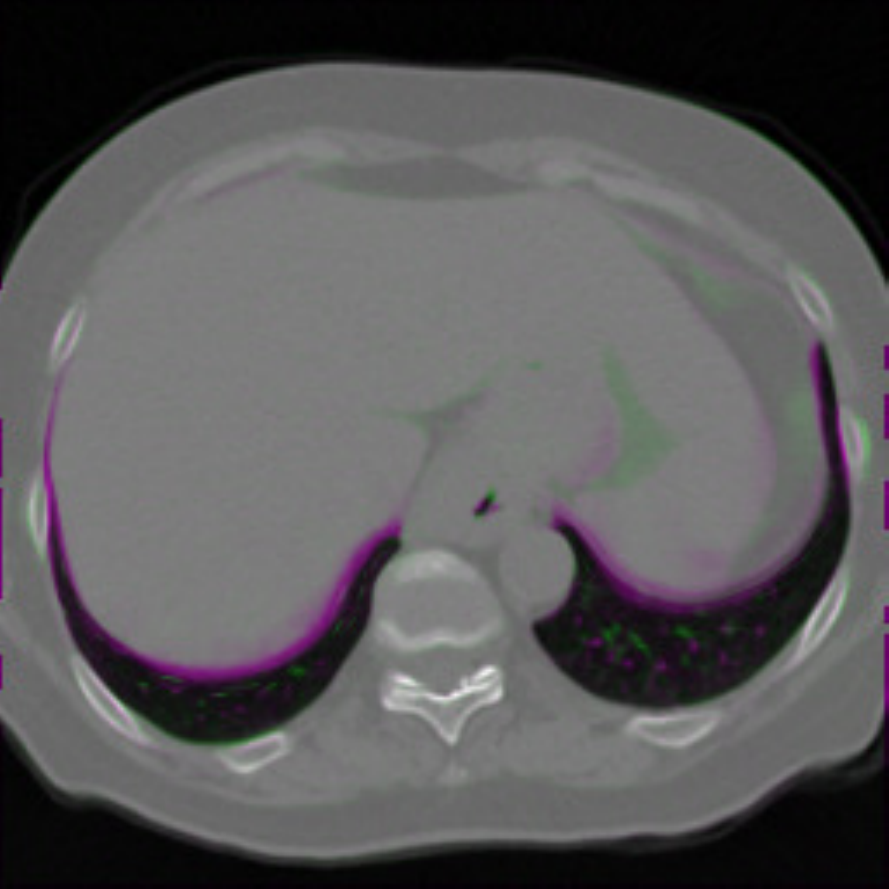}
\end{minipage}
\begin{minipage}[b]{0.133\linewidth}
\includegraphics[trim=1.0cm 1.4cm 0.0cm 0.00cm, clip=true,scale=0.21]{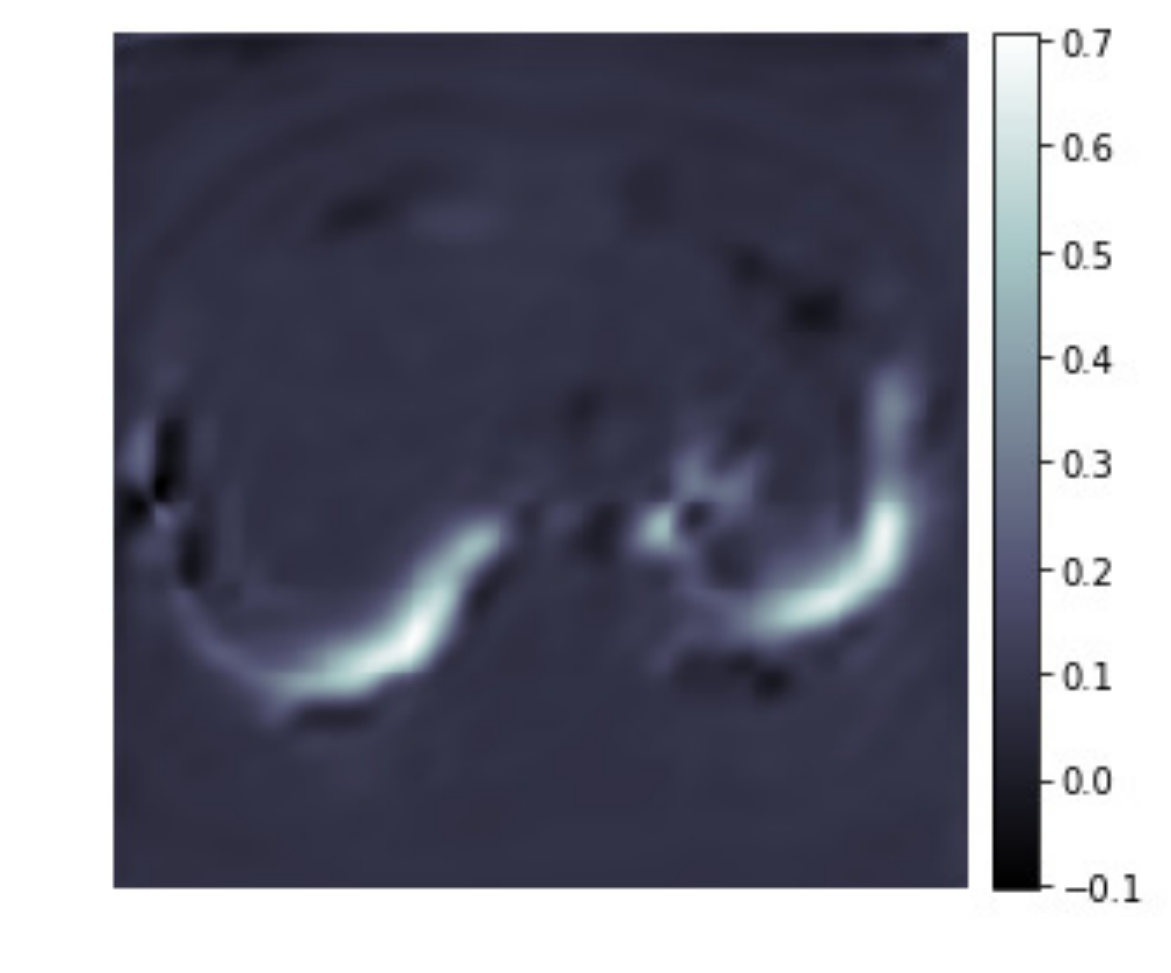}
\end{minipage}
\begin{minipage}[b]{0.133\linewidth}
\includegraphics[trim=1.0cm 1.4cm 0.0cm 0.00cm, clip=true,scale=0.21]{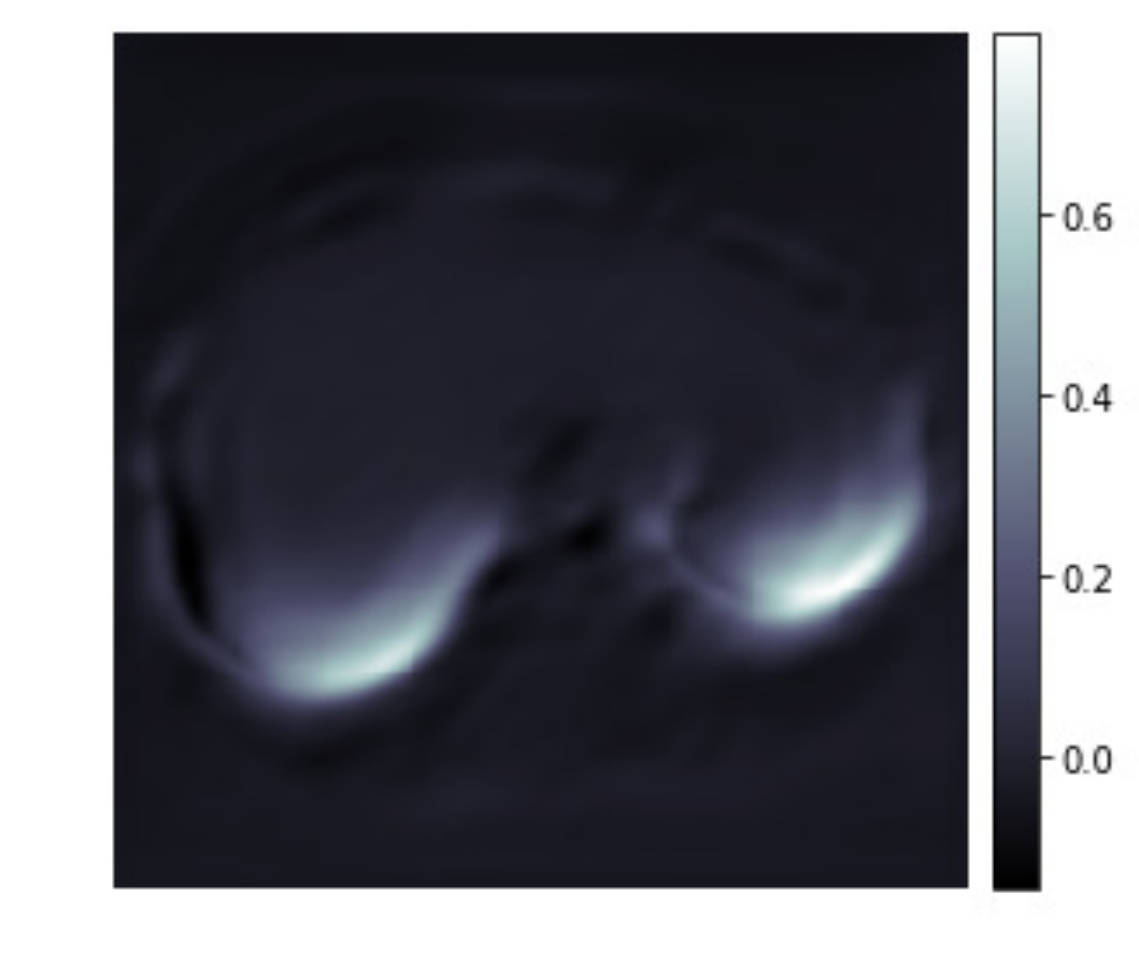}
\end{minipage}
\begin{minipage}[b]{0.135\linewidth}
\includegraphics[scale=0.19]{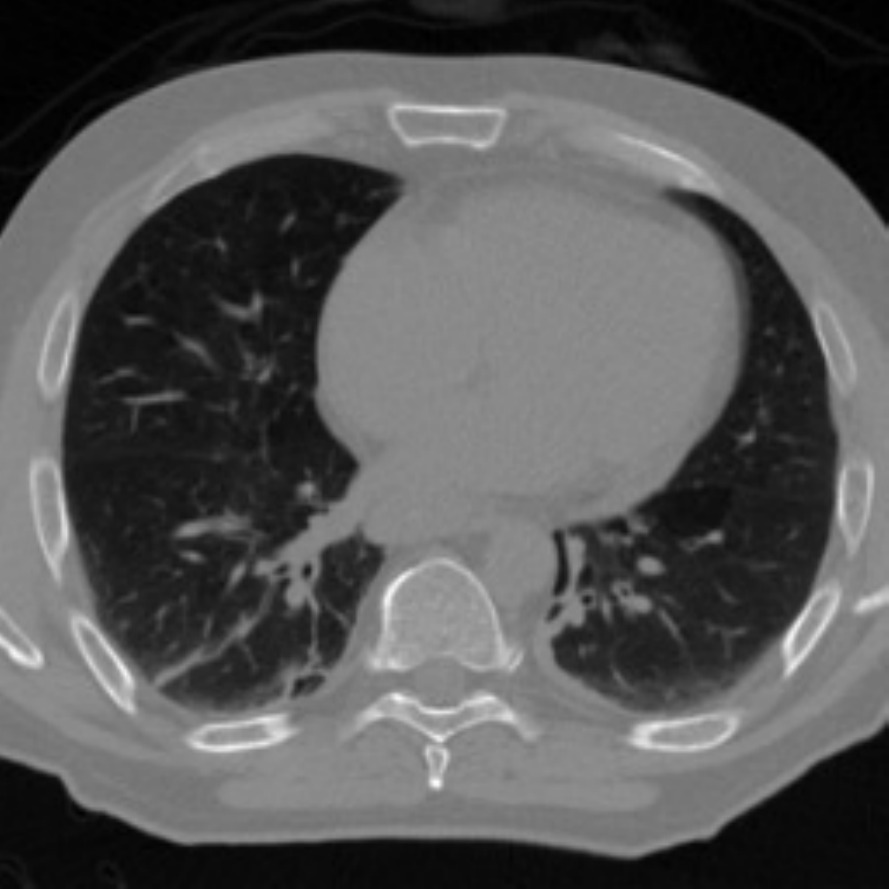}
\end{minipage}
\begin{minipage}[b]{0.135\linewidth}
\includegraphics[scale=0.19]{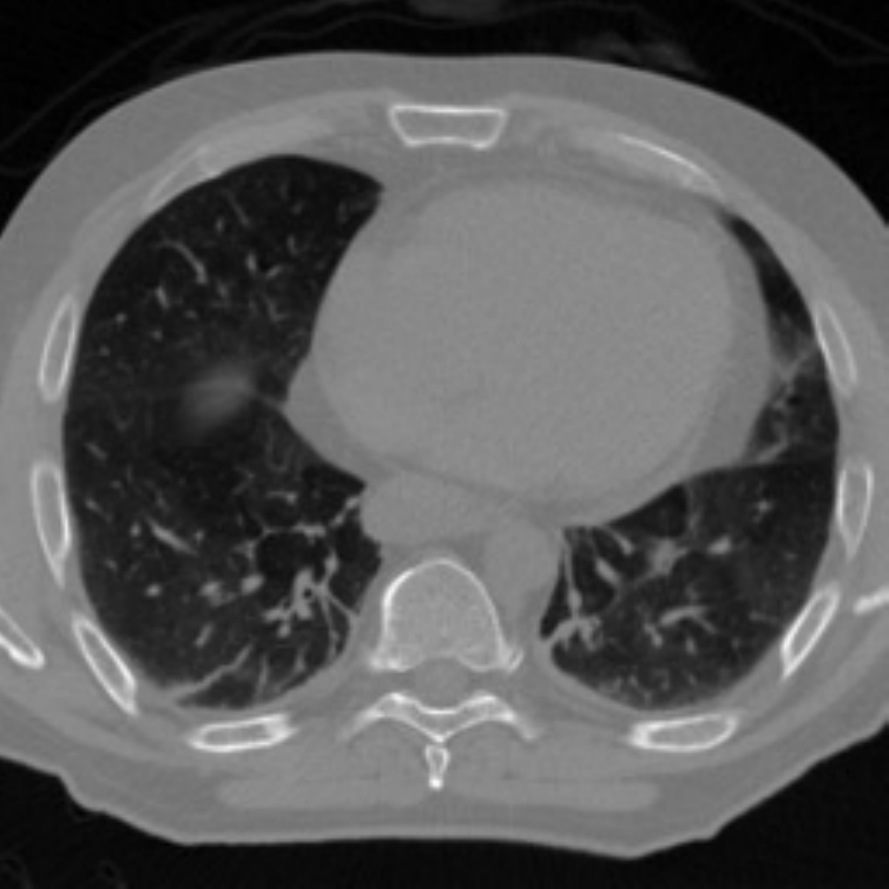}
\end{minipage}
\begin{minipage}[b]{0.135\linewidth}
\includegraphics[scale=0.19]{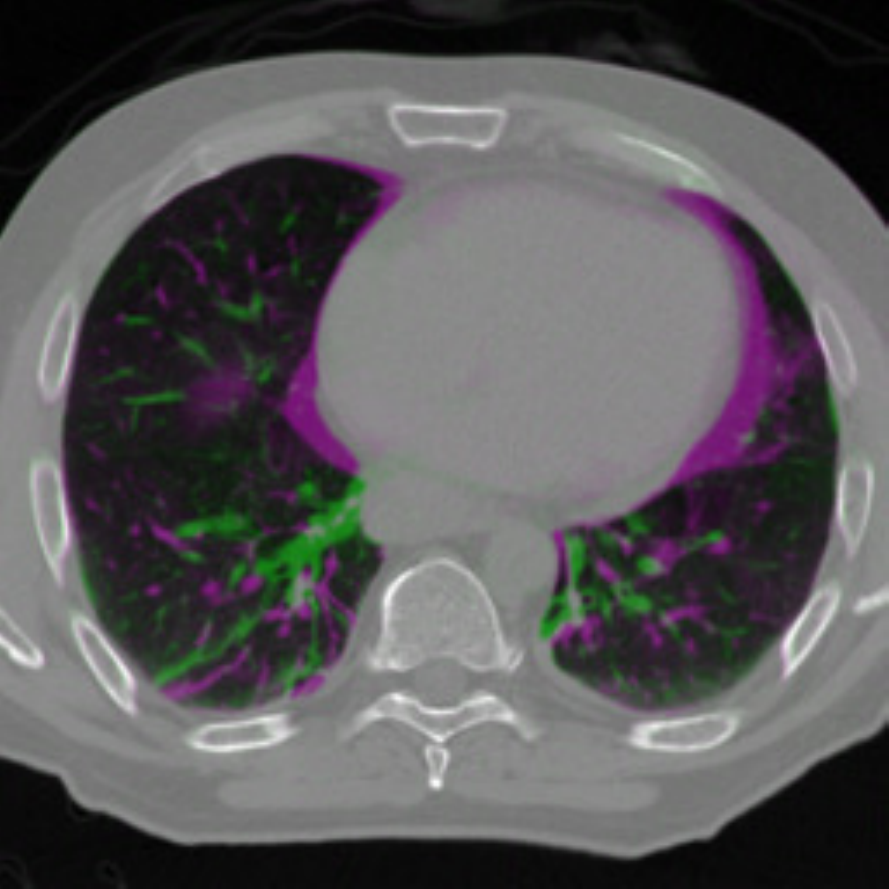}
\end{minipage}
\begin{minipage}[b]{0.135\linewidth}
\includegraphics[scale=0.19]{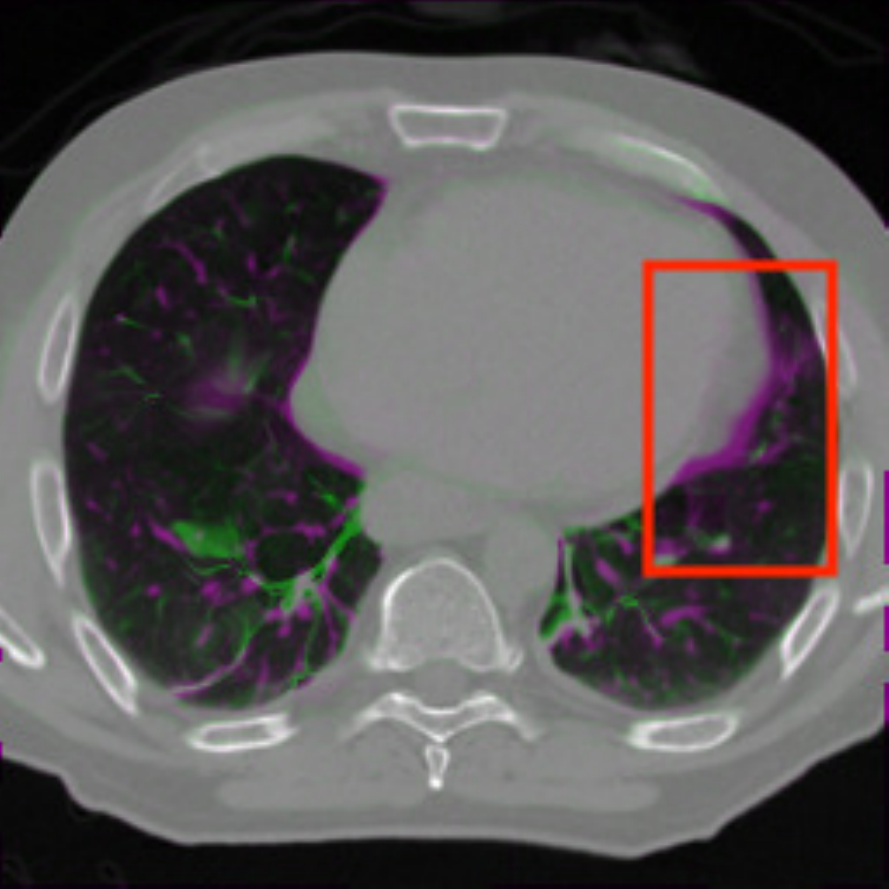}
\end{minipage}
\begin{minipage}[b]{0.135\linewidth}
\includegraphics[scale=0.19]{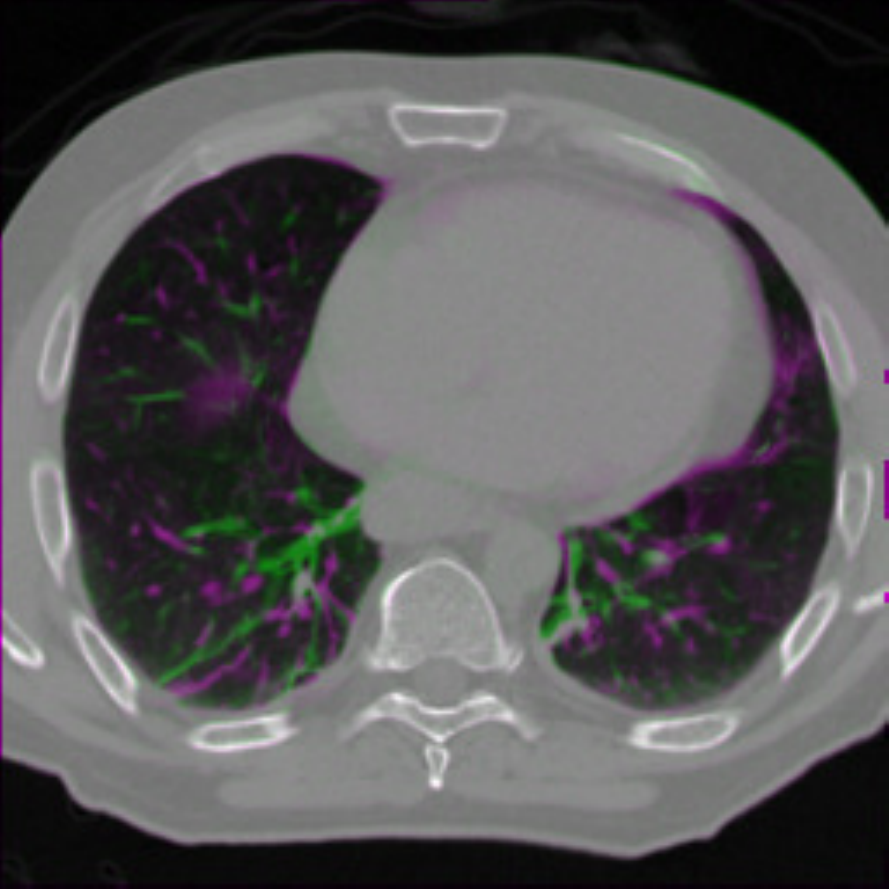}
\end{minipage}
\begin{minipage}[b]{0.135\linewidth}
\includegraphics[trim=1.0cm 1.4cm 0.05cm 0.05cm, clip=true,scale=0.21]{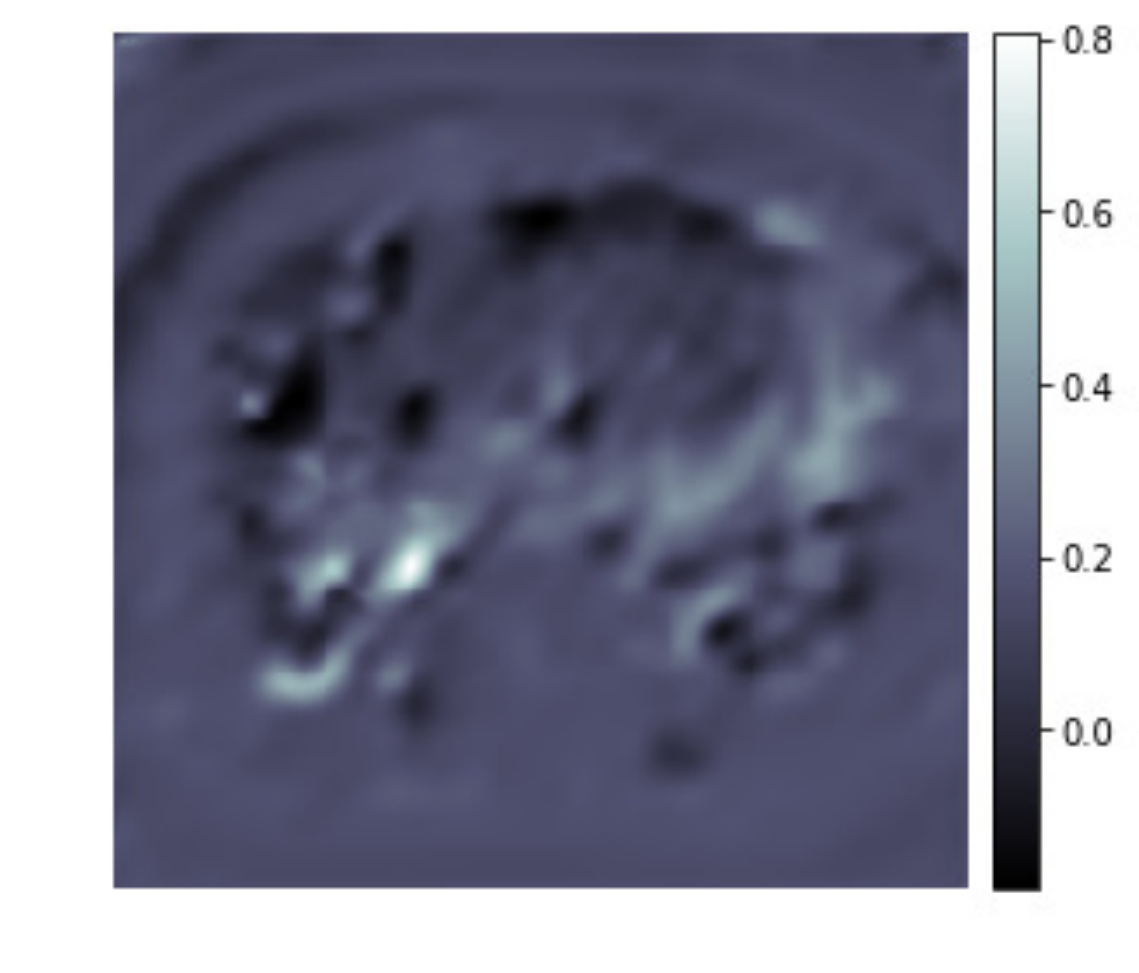}
\end{minipage}
\begin{minipage}[b]{0.135\linewidth}
\includegraphics[trim=1.0cm 1.5cm 0.0cm 0.05cm, clip=true,scale=0.21]{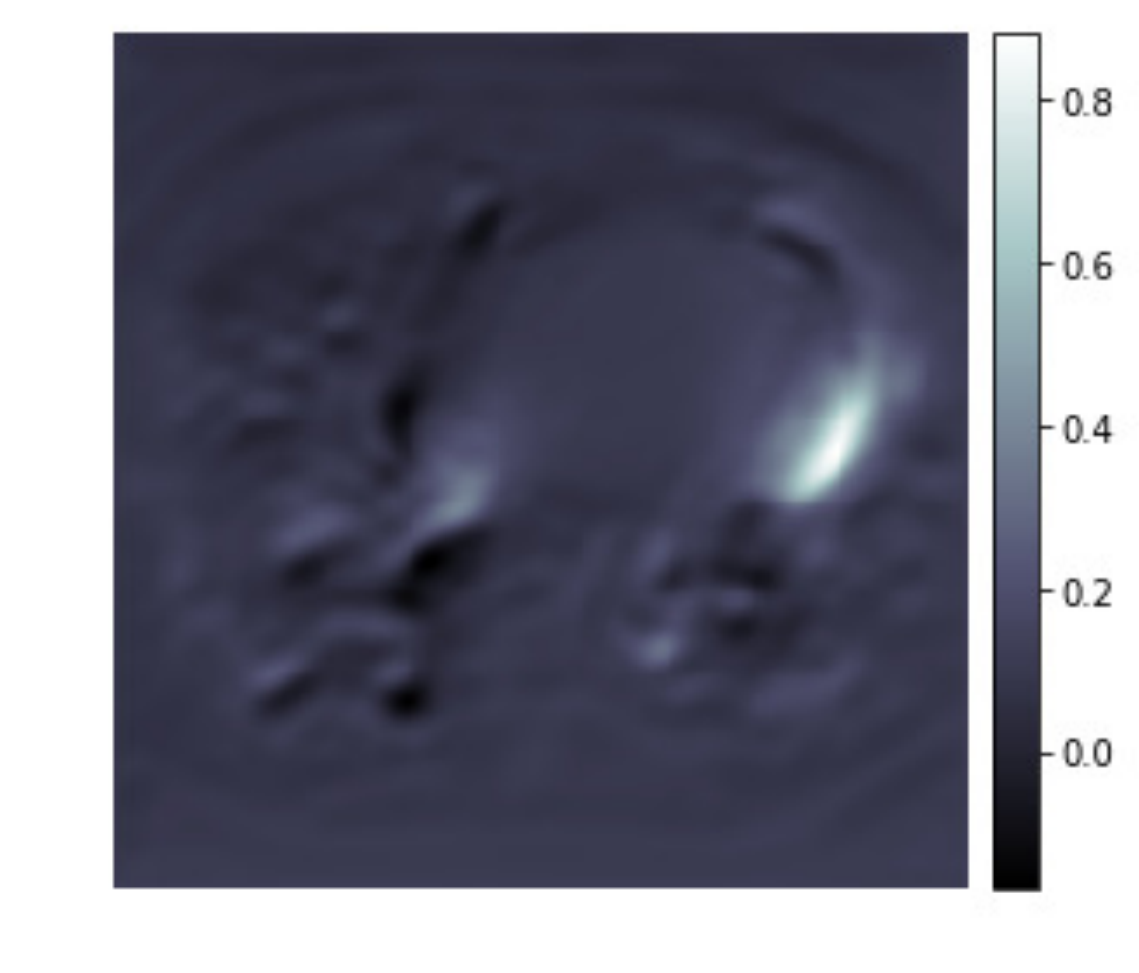}
\end{minipage}
\begin{minipage}[b]{0.135\linewidth}
\vspace{-20.00mm}
\centerline{${I}^{s}$}\medskip
\end{minipage}
\begin{minipage}[b]{0.135\linewidth}
\vspace{-20.00mm}
\centerline{${I}^{t}$}\medskip
\end{minipage}
\begin{minipage}[b]{0.135\linewidth}
\vspace{-20.00mm}
\centerline{$I^s\rightarrow{I}^t$}\medskip
\end{minipage}
\begin{minipage}[b]{0.135\linewidth}
\vspace{-5.00mm}
\centerline{${I}^{s}\small(\textbf{x}+\textbf{u}\small)$}\medskip
\end{minipage}
\begin{minipage}[b]{0.135\linewidth}
\vspace{-5.00mm}
\centerline{${I}^{s}(\textbf{x}+\textbf{u}_{reg})$}\medskip
\end{minipage}
\begin{minipage}[b]{0.135\linewidth}
\vspace{-0.50mm}
\centerline{$\mid\textbf{u}\mid$}\medskip
\end{minipage}
\begin{minipage}[b]{0.135\linewidth}
\centerline{$\mid\textbf{u}_{reg}\mid$}\medskip
\end{minipage}
\begin{minipage}[b]{0.135\linewidth}
\vspace{-0.50mm}
\centerline{}\medskip
\end{minipage}
\begin{minipage}[b]{0.135\linewidth}
\vspace{-0.50mm}
\centerline{}\medskip
\end{minipage}
\begin{minipage}[b]{0.135\linewidth}
\vspace{-50.50mm}
\centerline{overlay}\medskip
\end{minipage}
\begin{minipage}[b]{0.135\linewidth}
\vspace{-50.50mm}
\centerline{$+I^t$ overlay}\medskip
\end{minipage}
\begin{minipage}[b]{0.135\linewidth}
\vspace{-2.50mm}
\centerline{$+I^t$ overlay}\medskip
\end{minipage}
\begin{minipage}[b]{0.135\linewidth}
\centerline{}\medskip
\end{minipage}
\begin{minipage}[b]{0.13\linewidth}
\centerline{}\medskip
\end{minipage}
\vspace{-5mm}
\caption{{\bf{Visual validation of proposed Conv2warp method}} without ($\lambda=0$ ) and with regularization ($\lambda=0.001$ ) in our loss function on POPI dataset. Overlay images represents source or warped images in magenta and target image in green. Red rectangles in overlay images with ${I}^{s}(\textbf{x}+\textbf{u})$ show unrestricted flow of pixels when no regularization is used. Smoother deformation fields are obtained with regularized Conv2warp. Brighter pixels in magnitude images on left represent large displacements. \label{fig:qualititativeResults}}
\end{figure}
\begin{figure}[t!]
\centering
\includegraphics[trim=0.0cm 0.0cm 0.0cm 0.0cm, clip=true,scale=0.27]{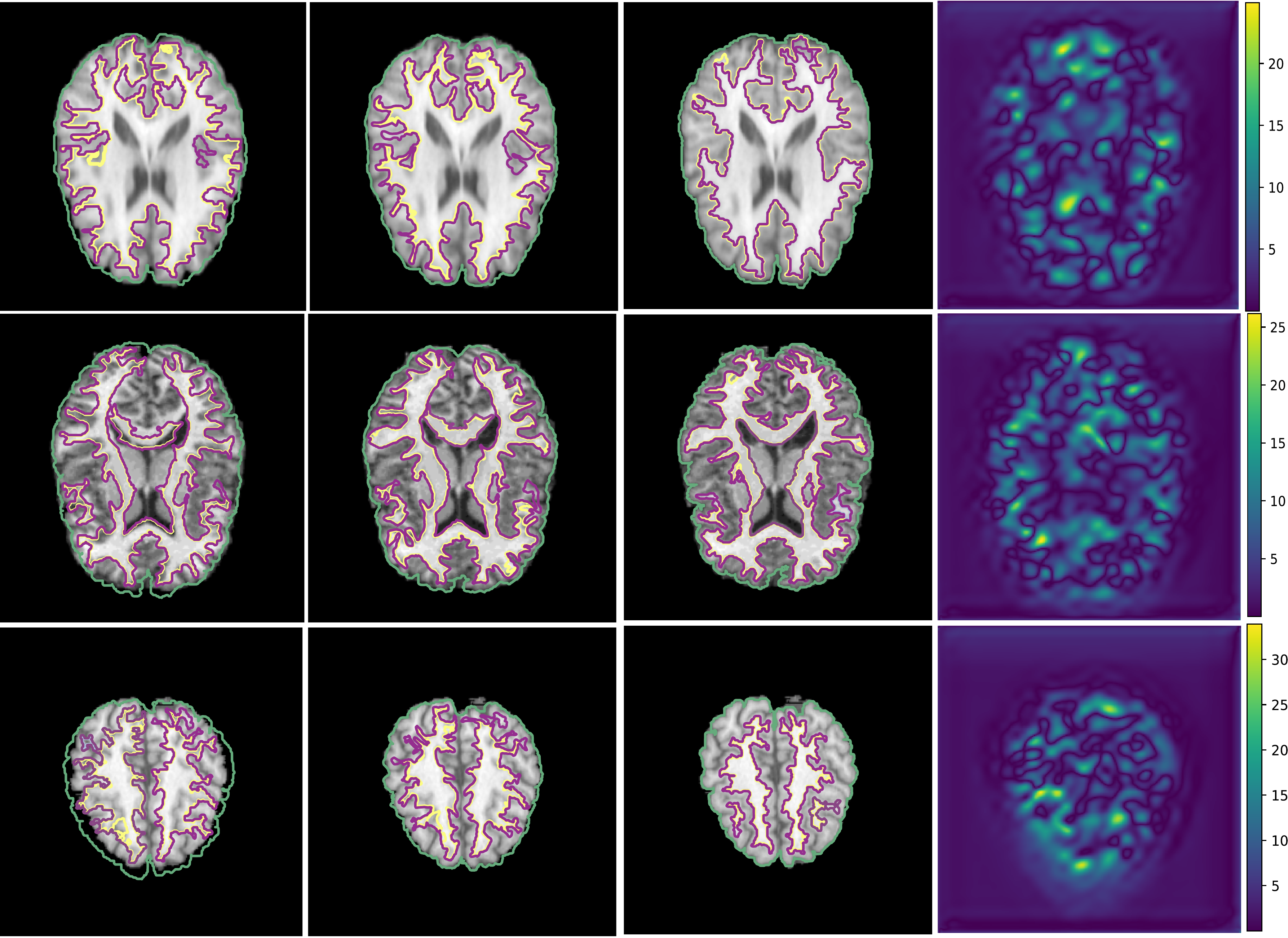}
\caption{{\bf{Visual validation of proposed Conv2Warp method on brain dataset.}} $1^{st}$ column: Target image with source mask, $2^{nd}$ column: target image with deformed (transformed) mask, $3^{rd}$ column: Target image with target mask, and last column: magnitude of predicted deformation field.  Grey matter (GM) and white-matter (WM) areas are marked by green and magenta borders and yellow borders, respectively.{\label{fig:qualitative2}}}
\end{figure}
\subsection{Registration of 4D CT data}
Conv2Warp has been evaluated by comparing it to the DIR state-of-the-methods. We also provide results where we restrict the convolution blocks in the Conv2Warp model to linear convolution only (ConvNet). For both these architectures, we will compare them only for Catmull-Rom spline based interpolation as it outperformed other spline techniques in our experiments. Seven pairs of different breathing cycles of 4D-CT data~\cite{POPI11}, each with 141 slices, where improvement of DIR methods compared to pre-alignment were significant are shown in Table~\ref{tab:POPI}. While the accuracy of our method measured by Dice (.90) and Jaccaard coefficient (.84) is similar to the rigorous simpleElastix (SE\footnote{http://simpleelastix.github.io}), the run time is reduced by a factor of 141. Conv2Warp outperforms all other state-of-the art methods and ConvNet for almost all considered pairs. The test time on CPU is nearly 2.94~s which is multiple-folds lower than conventional DIR methods. Our model is light weight with inference time on a GPU of less than 1~s.
\subsection{Registration of T1 MRI data}
T1-weighted MRI scans for 9 volume pairs of MGH10 and 11 volume pairs of CUMC12~\cite{Klein:NI09} where each image from different volumes were registered to the first volume data were also used for evaluation. The weights trained on LBPA~\cite{LBPA40} dataset were used in this case. In Table~\ref{tab:BRAIN} Conv2Warp have $\mu_{dice}$ (mean dice) and $\mu_{jaccard}$ (mean Jaccard) of 0.95 and 0.90, respectively for MGH10 dataset (same as for SE), and 0.97 and 0.93, respectively for CUMC12 dataset (higher than SE). Conv2Warp has higher $\mu_{dice}$ and $\mu_{jaccard}$ for both datasets compared to ANTS (SyN-CC) and ConvNet. When compared with speed Conv2Warp is computationally the fastest among all other conventional DIR methods.
%%%%%%%%%%%%%%
\subsection{Visual validation}
%%%%%%%%%%%%%%
In Fig.~\ref{fig:qualititativeResults} Conv2Warp without ($\lambda=0$) regularization and with ($\lambda=0.001$) regularization are shown in 4th and 5th columns respectively and their corresponding DVF magnitudes in 6th and 7th columns, respectively. It can be observed that the unconstrained loss function results in some unrealistic deformations (red rectangular regions in 4th column) while a more realistic deformations are visible for the constrained loss proposed in Conv2Warp (5th column). A smooth deformation can be seen in the magnitude image of the DVF $\mid \textbf{u}_{reg} \mid$ (7th column). Colour overlay images show a large improvement in the alignment of source images $I^s$ with the target images $I^t$ (3rd column) with Conv2Warp method (5th column). Fig.~\ref{fig:qualitative2} shows the results on T1 MRI test datasets which were first rigidly aligned to MNIspace and then co-registered using Conv2Warp. It can be observed that Conv2Warp handles different magnitudes of non-linear deformations.
% present in this data as well.  
%%%%%%%%%%%%%%Additional qualitative results are in Suppl. Mat. Section 3.
\section{Conclusion}
\label{sec:conc}
We have proposed a novel end-to-end convolutional neural network that consists of a sequential linear and deformable convolutions along with a learnt non-linear sampler. To handle wide range of non-linear deformations between source and target data pairs deformations are concurred by the continuous warping strategy. Our experimental results demonstrate that our proposed model outperforms most of the traditional methods and has very low computational complexity. When compared to the existing linear deep learning models such as ConvNets, Conv2Warp produces more accurate deformation fields. Additional details and evaluations are available in the supplementary material. 
\subsubsection*{Acknowledgments}
SA is supported by the NIHR Oxford BRC and JR by EPSRC EP/M013774/1 Seebibyte.
\bibliographystyle{splncs03}
\bibliography{registration}
\end{document}